\documentclass[11pt, a4paper]{article}      
\usepackage{amsfonts}
\usepackage[urlcolor=blue]{hyperref}
\newif\ifpdf
\ifx\pdfoutput\undefined
\pdffalse 
\else
\pdfoutput=1 
\pdftrue
\fi

\ifpdf
\usepackage[pdftex]{graphicx}
\else
\usepackage{graphicx}
\fi

\title{Conceptors: an easy introduction}

\author{Herbert Jaeger, Jacobs University Bremen}

\begin{document}

\maketitle 

\begin{abstract} Conceptors provide an elementary
  neuro-computational mechanism which sheds a fresh and unifying light
  on a diversity of cognitive phenomena.  A number of demanding
  learning and processing tasks can be solved with unprecedented ease,
  robustness and accuracy. Some of these tasks were impossible to
  solve before. This entirely informal paper introduces the basic
  principles of conceptors and highlights some of their usages.
\end{abstract}

\section{The big picture}

The subjective experience of a functioning brain is wholeness: \emph{I}! 

Scientific analysis explodes this unity into a myriad of phenomena,
functions, mechanisms and objects: \emph{abstraction, action, action potential,
  actuator, adaptation, adult, affect, aging, algorithm, amygdala, ...}:
just a quick  pick from the  subject indices of
psychology, neuroscience, machine learning, AI, cognitive science,
robotics, linguistics, psychiatry.

How to re-integrate these scattered items into  functioning whole
from which they sprang?

Again and again, integrative views of brains and cognition were advanced:
behaviorism; the cybernetic brain; hyperstability; general
problem solver; physical symbol systems; society of mind; synergetics;
autopoietic systems; behavior-based agents; ideomotor theory; the
Bayesian brain. Yet, the very multiplicity of such paradigms attests
to the  perpetuity of the integration challenge.


Conceptors offer novel options to take us a few concrete steps
further down the long and winding road to cognitive system
integration.

Conceptors are a neuro-computational mechanism which -- basic and
generic like a stem cell -- can differentiate into a diversity of
neuro-computational functionalities: incremental learning of dynamical
patterns; perceptual focussing; neural noise suppression;
morphable motor pattern generation; generalizing from a few learnt
prototype patterns; top-down attention control in hierarchical online
dynamical pattern recognition; Boolean combination of evidence in
pattern classification; content-addressable dynamical pattern memory;
pointer-addressable dynamical pattern memory (all demonstrated by
simulations in \cite{Jaeger14extended}).  In this way they suggest a common
computational principle underneath a number of seemingly diverse
neuro-cognitive phenomena.

Conceptors can be formally or computationally instantiated in several
ways and on several levels of abstraction: as neural circuits, as
adaptive signal filters, as linear operators in dynamical systems, as
operands in an extended Boolean calculus, and as categorical objects
in a logical framework (all detailed in \cite{Jaeger14extended}). In this way
they establish new translation links between different scientific views,
in particular between numeric-dynamical and symbolic-logical accounts
of neural and cognitive processing.

\section{The basic mechanism}

Conceptors can be intuitively explained in three steps.

\begin{figure}[htb]
\center
  \includegraphics[width=100mm]{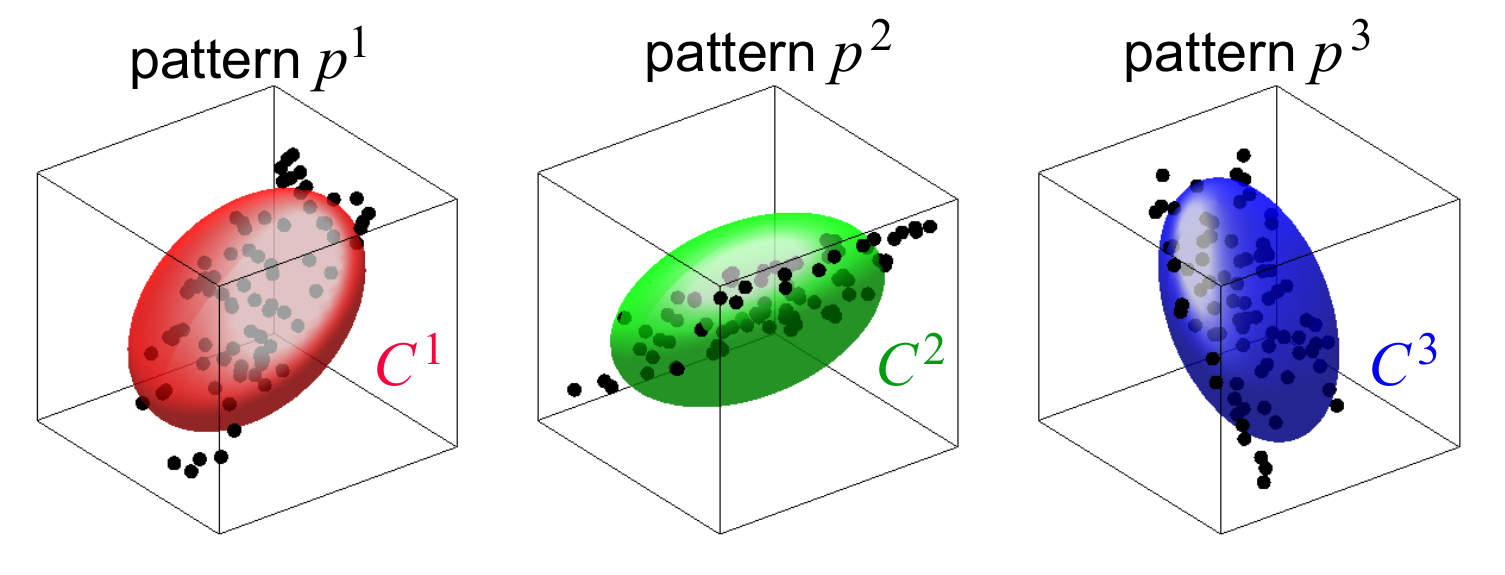}
\caption{Conceptor geometry (schematic, here network size $N =
  3$). Three patterns $p^1, p^2, p^3$ excite neural state  
  clouds (black dots) whose shapes can be characterized by ellipsoids
  (red, green, blue) corresponding to  conceptors $C^1, C^2,
  C^3$.}
\label{fig1}
\end{figure}

\emph{Step 1: From dynamical patterns to conceptors.} Consider a
recurrent neural network (RNN) $\mathcal{N}$ with $N$ neurons which is
driven by several dynamical input patterns $p^1, p^2, ...$ in turn.
The concrete type of RNN model (spiking or not, continuous or discrete
time, deterministic or stochastic) is of no concern, and the patterns
$p^j$ may be stationary or non-stationary, scalar or multidimensional
signals. When $\mathcal{N}$ is driven with pattern $p^j$, the
$N$-dimensional excited neural states $\{x^j\}$ come to lie in a state
cloud whose geometry is characteristic of the driving pattern. The
simplest formal characterization of this geometry of $\{x^j\}$ is
given by an ellipsoid $C^j$ whose main axes are the principal
components of the state set $\{x^j\}$ (Figure \ref{fig1}).  This
ellipsoid $C^j$ represents \emph{the conceptor associated with pattern
  $p^j$ in the network $\mathcal{N}$}.  $C^j$ can be concretely
instantiated in various ways, for instance as a matrix, as a separate
subnetwork, or as a single neuron projecting to a large random
``reservoir'' network. In any case, $C^j$ can be learnt from $\{x^j\}$
by a variety of simple and robust learning rules which all boil down
to the objective ``learn a regularized identity map''.

\begin{figure}[htb]
\center
  \includegraphics[width=100mm]{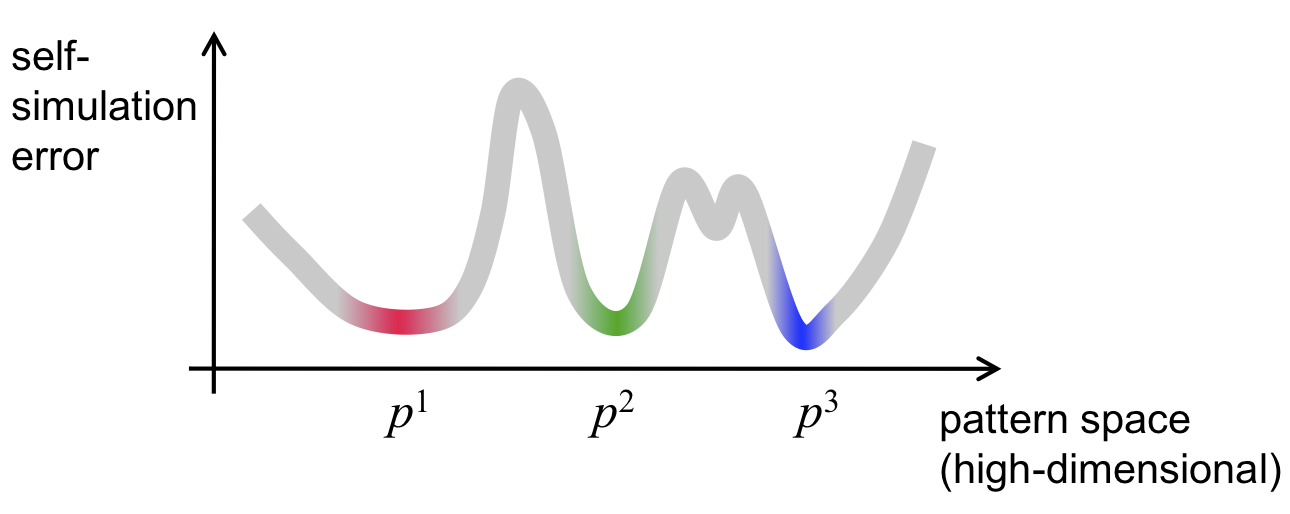}
\caption{Self-simulation error after storing three patterns $p^1, p^2,
  p^3$ (schematic). The trained network can replicate the untrained
  networks' driven dynamics with small error for the stored driving
  patterns. }
\label{fig2}
\end{figure}

\emph{Step 2: Storing prototype patterns.} In order to realize some of
the potential conceptor functionalities, the patterns $p^1, p^2, ...$
must be \emph{stored} in the RNN $\mathcal{N}$.  The objective
defining this storing task is that the network learns to replicate the
pattern-driven state sequences $\{x^j\}$ in the absence of the driver.
This could be called a ``self-simulation'' objective.  It can be
effected by an elementary RNN adaptation scheme which in the last few
years has been independently introduced under the names of
``self-prediction'' (Mayer \& Browne), ``equilibration'' (Jaeger),
``reservoir regularization'' (Reinhart \& Steil), ``self-sensing
networks'' (Sussillo \& Abbott), and ``innate training'' (Laje \&
Buonomano).  Write $\mathcal{N}(p^1, \ldots, p^n)$ for the network
obtained after $n$ patterns have been stored.  In intuitive terms one
could say that the storing procedure entrenches the various
pattern-driven dynamics $\{x^j\}$ (where $j = 1,\ldots,n$) into the
network (visualized in Figure \ref{fig2}).  However, these entrenched
dynamics are inherently unstable due to crosstalk. 

 \begin{figure}[htb]
\center
  \includegraphics[width=100mm]{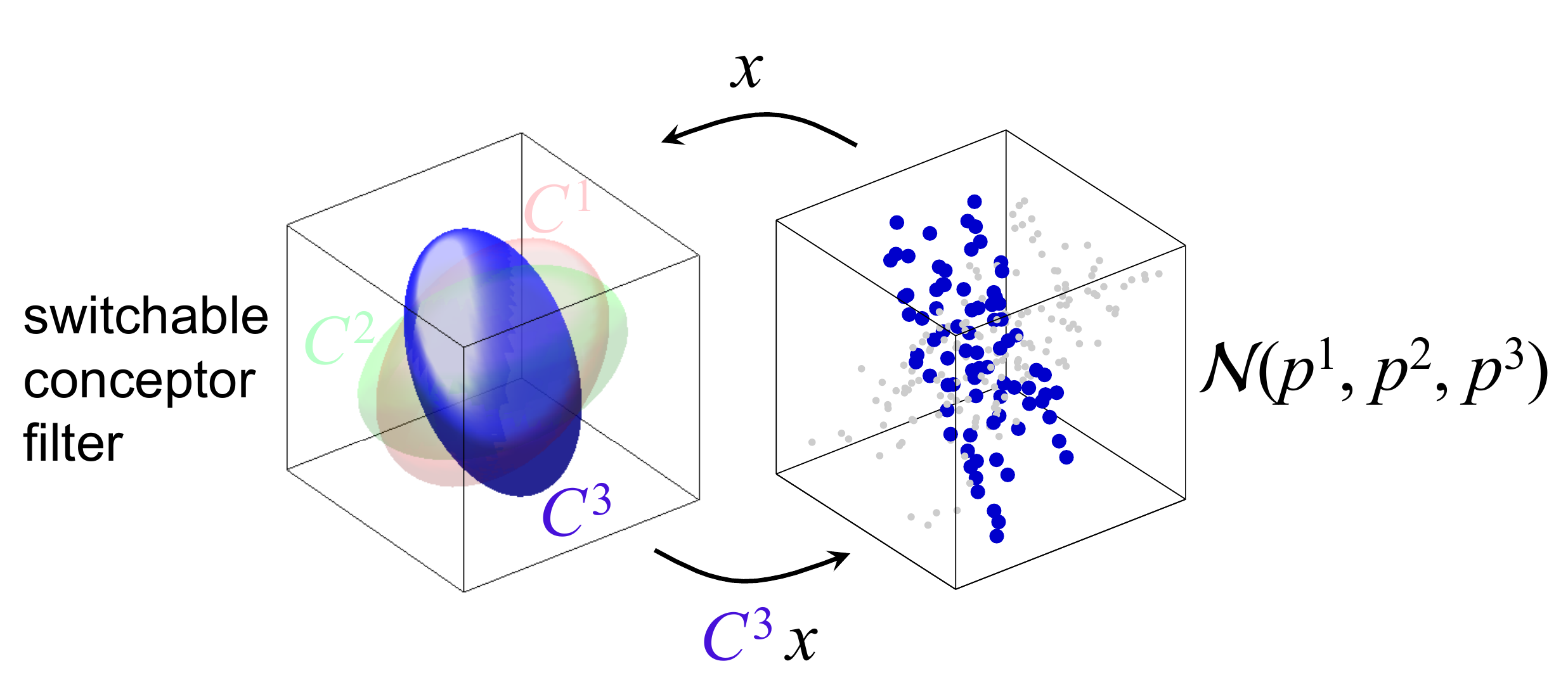}
\caption{Basic usage of conceptors. A switchable
  conceptor filter is inserted into the update loop of the RNN. In
  this schematic example the third (blue) of three patterns stored in
  $\mathcal{N}(p^2, p^2, p^3)$ is selected and stabilized by
  activating the associated conceptor $C^3$. }
\label{fig3}
\end{figure}

\emph{Step 3: From conceptors to dynamical patterns.} If
$\mathcal{N}(p^1, \ldots, p^n)$ were just let running freely (in the
absence of input), unpredictable behavior would result due to the
instability of the entrenched dynamics. Now the conceptors
$C^1,\ldots, C^n$ associated with $p^1,\ldots, p^n$ are called on
stage. Assume we want the network to re-generate the dynamics
$\{x^j\}$ associated with pattern $p^j$ - stably and accurately. This
is achieved by inserting the corresponding conceptor $C^j$ into the
recurrent update loop of  $\mathcal{N}(p^1, \ldots, p^n)$.  In
mathematical abstraction, $C^j$ is a linear map and inserting it means
to insert the operation $x \mapsto C^j\,x$ into the recurrent 
state update loop.  In intuitive geometric terms, the network states
are \emph{filtered} by the ellispoid shape of $C^j$: state components
aligned with the ``thick'' dimensions of this ellipsoid pass
essentially unaltered whereas components in the ``flat'' directions
are suppressed (Figure \ref{fig3}). As a result, the neural dynamics
$\{x^j\}$ corresponding to pattern $p^j$ is \emph{selected} and
\emph{stabilized}. Changing from one conceptor $C^j$ to another
conceptor $C^i$ swiftly switches network dynamics from one mode to the
next. The ``insertion'' can be mathematically, biologically or
technically implemented in various ways, depending on how $C^j$ is
concretely instantiated. Implementation options range from
matrix-based conceptor filters (convenient in machine learning
applications) to activating neurons which represent conceptors (in
biologically more plausible ``reservoir networks'' realizations of
conceptors) \cite{Jaeger14extended}.
\vspace{0.5cm}

\noindent \emph{Summary -- the essence of conceptor mechanisms:}
\begin{enumerate}
\item Different driving patterns lead to differently shaped state
clouds in a driven RNN. The ellipsoid envelopes of these clouds make
conceptors.  
  \item After driving patterns have been stored in the network, they
can be selected and stably re-generated  by inserting the
corresponding conceptor filters in the update loop.
\end{enumerate}

\section{A little more detail and some highlight examples}

 \begin{figure}[htb]
\center
  \includegraphics[width=70mm]{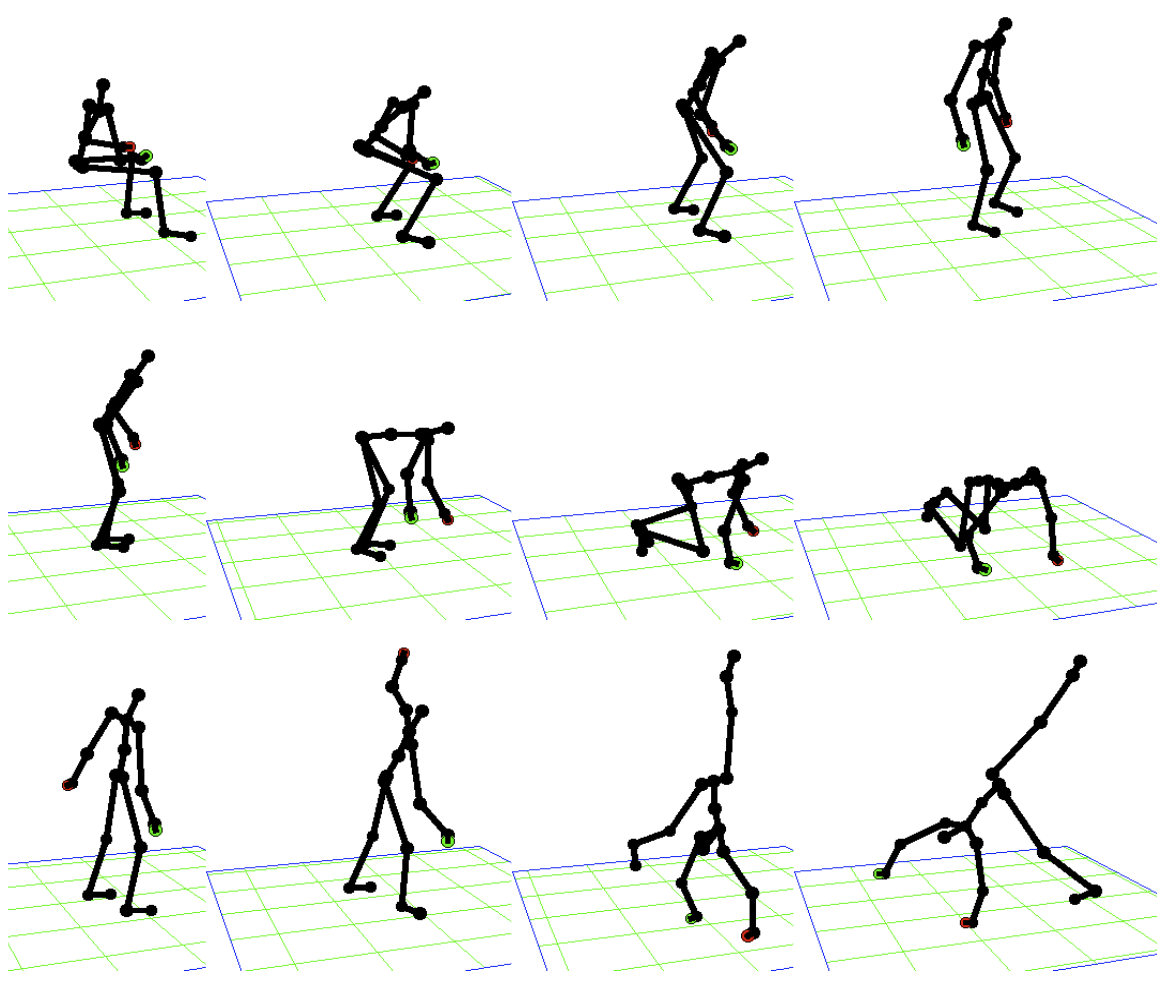}
\caption{Impressions from a human motion sequence generated by an RNN
  under conceptor control. Watch video:
  \href{http://youtu.be/DkS_Yw1ldD4}{\tt youtu.be/DkS\_Yw1ldD4}.}
\label{fig4}
\end{figure}

The most basic use case for conceptors is to store a number of
patterns in an RNN and later re-play them: a neural long-term memory
for dynamical patterns with addressing by conceptors. Demo: 15 human
motion patterns were stored in a 1000-neuron RNN. These patterns were
61-dimensional signals distilled from human motion capture data
retrieved from the CMU mocap repository
(\href{http://mocap.cs.cmu.edu}{\tt mocap.cs.cmu.edu}). Some of these
patterns were periodic, others were transient. A single short training
sequence per pattern was used for training. In
order to re-generate a composite motion sequence from the network,
associated conceptors were activated in turn and the obtained network
dynamics was visualized (using the mocap visualization
toolbox from the University of Jyv\"{a}skyl\"{a}
\href{http://www.jyu.fi/hum/laitokset/musiikki/en/research/coe/materials/mocaptoolbox}{\tt
  www.jyu.fi/hum/laitokset/musiikki/en/research/coe/materials/
  \\mocaptoolbox }). Figure \ref{fig4} shows some thumbnails. Smooth
transitions between successive motion patterns $p^i, p^j$ were
obtained by linearly blending the conceptor matrix $C^i$ into $C^j$
for a one simulated second.

\begin{figure}[htb]
\center
  \includegraphics[width=70mm]{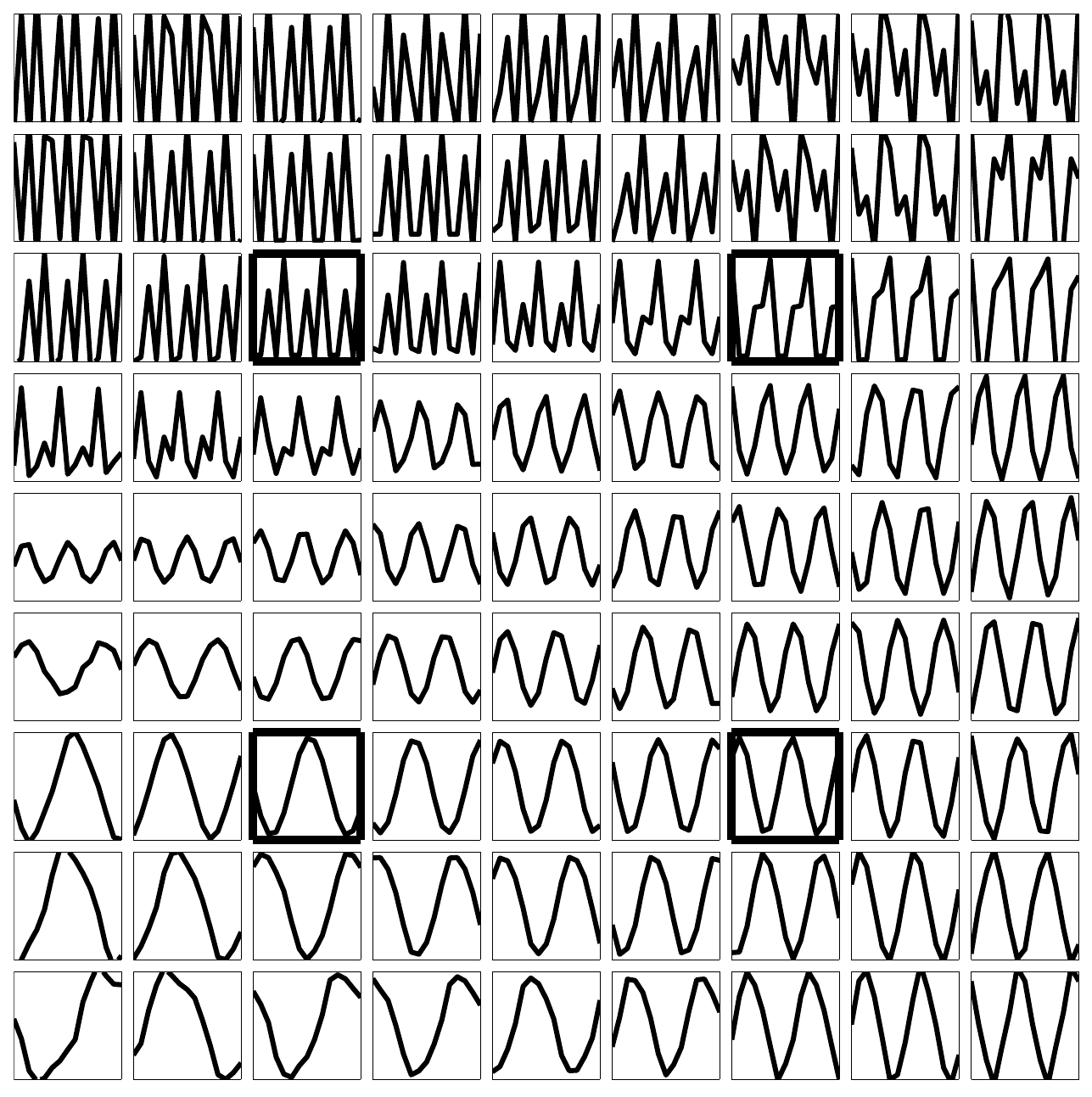}
\caption{Morphing patterns by mixing conceptors.}
\label{fig5}
\end{figure}

When some ``prototype'' patterns have been stored, they can be \emph{morphed}
in recall by using linear mixes of the prototype conceptors. Demo:
four patterns $p^1,\ldots,p^4$ were stored, two of which were
5-periodic random patterns and the the other two were sampled
irrational-period sines. Figure \ref{fig5} shows the result of using
conceptor mixtures $a_1 C^1 + \ldots + a_4 C^4$, where $\sum a_i = 1$.
When all mixing coefficients are nonnegative one obtains an
interpolation between, when some are negative one gets extrapolation
beyond the stored prototypes. The four panels with bold outlines show
the recalled prototypes (one of the $a_i$ is equal to 1, the others
are 0). Note that interpolations are created even between
integer-periodic and irrational-period signals, which in the
terminology of dynamical systems correspond to attractors of
incommensurable topology (point attractors vs. attractors with the
topology of the unit circle).

\begin{figure}[htb]
\center
  \includegraphics[width=100mm]{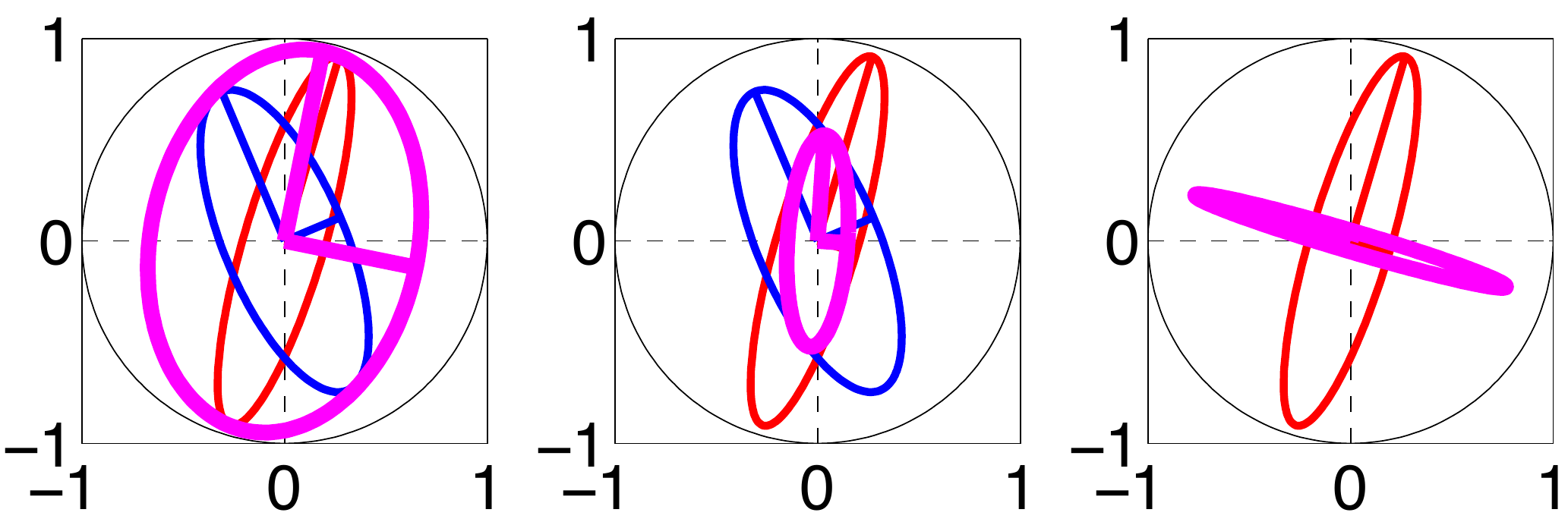}
\caption{Geometry of Boolean operations on conceptors (2-dimensional
  case shown). From left to right: OR, AND, NOT. Thin blue and red
  ellipses: arguments, thick magenta ellipse: result of the respective
  operation.}
\label{fig6}
\end{figure}

Conceptors can be combined by operations OR (written $\vee$), AND
($\wedge$), NOT ($\neg$) which obey
almost all laws of Boolean logic (for certain classes of conceptors,
full Boolean logic applies) and which admit a rigorous semantical
interpretation. For instance, the OR of two conceptors $C^i, C^j$,
which are individually derived from neural state sets $\{x^i\},
\{x^j\}$, is (up to a normalization scaling) the conceptor that would be
derived from the union of these two state sets. Figure \ref{fig6}
illustrates the geometry of these operations. 

These Boolean operations furthermore induce an \emph{abstraction}
ordering $\leq$ on conceptors by defining $A \leq B$ if there exists
some $C$ such that $B = A \vee C$. When conceptors are represented as
matrices, this logical definition of $\leq$
coincides with the  well-known L\"{o}wner
ordering of matrices. When conceptors are represented as certain adaptive
neural circuits, deciding whether $A \leq B$ computationally amounts
to checking whether the activation of certain neurons increases. The
extreme cheapness of this local check may give a hint why humans can often
make classification judgements with so little apparent effort.

\begin{figure}[htb]
\center
  \includegraphics[width=150mm]{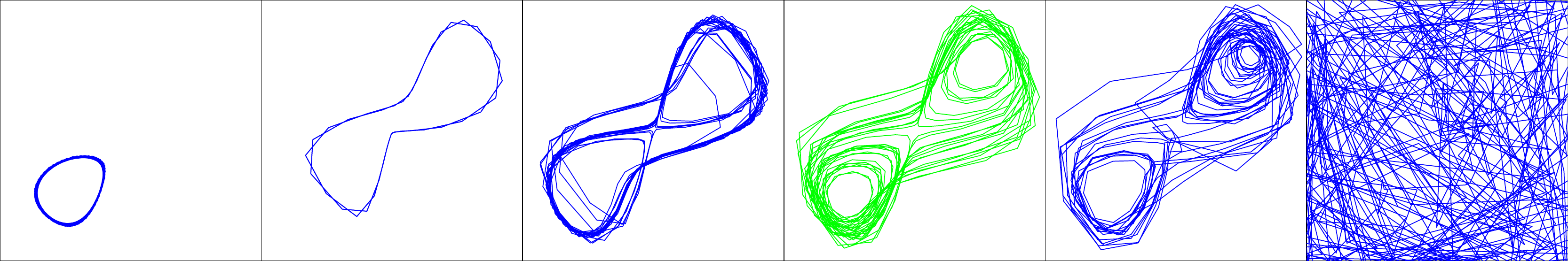}
\caption{Opening the aperture of a conceptor. The Lorenz chaotic attractor
  pattern is re-generated with its conceptor set at different levels
  of aperture. Panels show delay-embedded plots of a scalar observer
  of the regenerated patterns. From left to right: aperture is opened
  up. The panel drawn in green marks an aperture value (found
  automatically) where the stored Lorenz pattern is re-generated with
  high precision.}
\label{fig7}
\end{figure}

A special case of abstraction is defocussing. In geometric terms, a
conceptor becomes defocussed if its ellipsoid shape is inflated by
a certain scaling operation which is governed by a parameter called
\emph{aperture}. At zero aperture a conceptor contracts to the zero
mapping, while when the aperture grows to infinity the conceptor
approaches the identity mapping. The larger the aperture, the more
signal components may pass through the state filtering $x \mapsto Cx$.
Demo: four different chaotic attractor patterns were stored in an RNN,
one of them being the well-known Lorenz attractor. When the conceptor
corresponding to the Lorenz attractor is applied at increasing
aperture levels, the re-generated pattern first goes through stages of
increasing differentiation, then in a certain aperture range becomes a
faithful replica of that attractor, after which it gradually becomes
over-excited and at very large aperture dissolves into the entirely
unconstrained behavior of the native network (Figure \ref{fig7}). An
optimal aperture can be autonomously adjusted by the system,
exploiting a cheaply measurable ``auto-focussing'' criterion based on
the signal damping ratio imposed by the conceptor.

With the help of Boolean operations and the abstraction ordering, a
network's conceptor repertoire can be viewed as being organized in an
abstraction hierarchy which shares many formal properties with
semantic networks and ontologies known from AI, linguistics and
cognitive science. This line of analysis can be extended to a full
account of conceptor systems in the modern category-theoretic setting
of logical frameworks, establishing a rigorous link between neural
dynamics and symbolic logic \cite{Jaeger14extended}.

Besides such uses for a scientific analysis, Boolean operations offer concrete 
computational exploits. One of them is incremental (life-long)
learning of dynamical patterns. The objective here is to store more
and more patterns in a network such that patterns stored later do not
catastrophically interfere with previously acquired ones. Let $p^1,
p^2, \ldots$ be a potentially open-ended series of patterns with
associated conceptors $C^1, C^2,\ldots$. In informal terms,
incremental storing can be achieved as follows. Assume the first $n$
patterns have been stored, yielding $\mathcal{N}(p^1, \ldots, p^n)$.
Characterize the neural memory space claimed by these patterns by $A^n
= C^1 \vee \ldots \vee C^n$ and the still free memory space by $F^n =
\neg A^n$. The next pattern $p^{n+1}$ with its conceptor $C^{n+1}$
typically has some dynamical components that are shared with some of
the already stored patterns, and it will have some  new dynamical
components. The latter can be characterized by the conceptor $N^{n+1}
= C^{n+1} - A^n$ (logical difference operator). The storing procedure
can be straightforwardly modified such that only the new dynamical
components characterized by $N^{n+1}$ are stored into the still
free memory space $F^n$.

     \begin{figure}[htb]
\center
  \includegraphics[width=130mm]{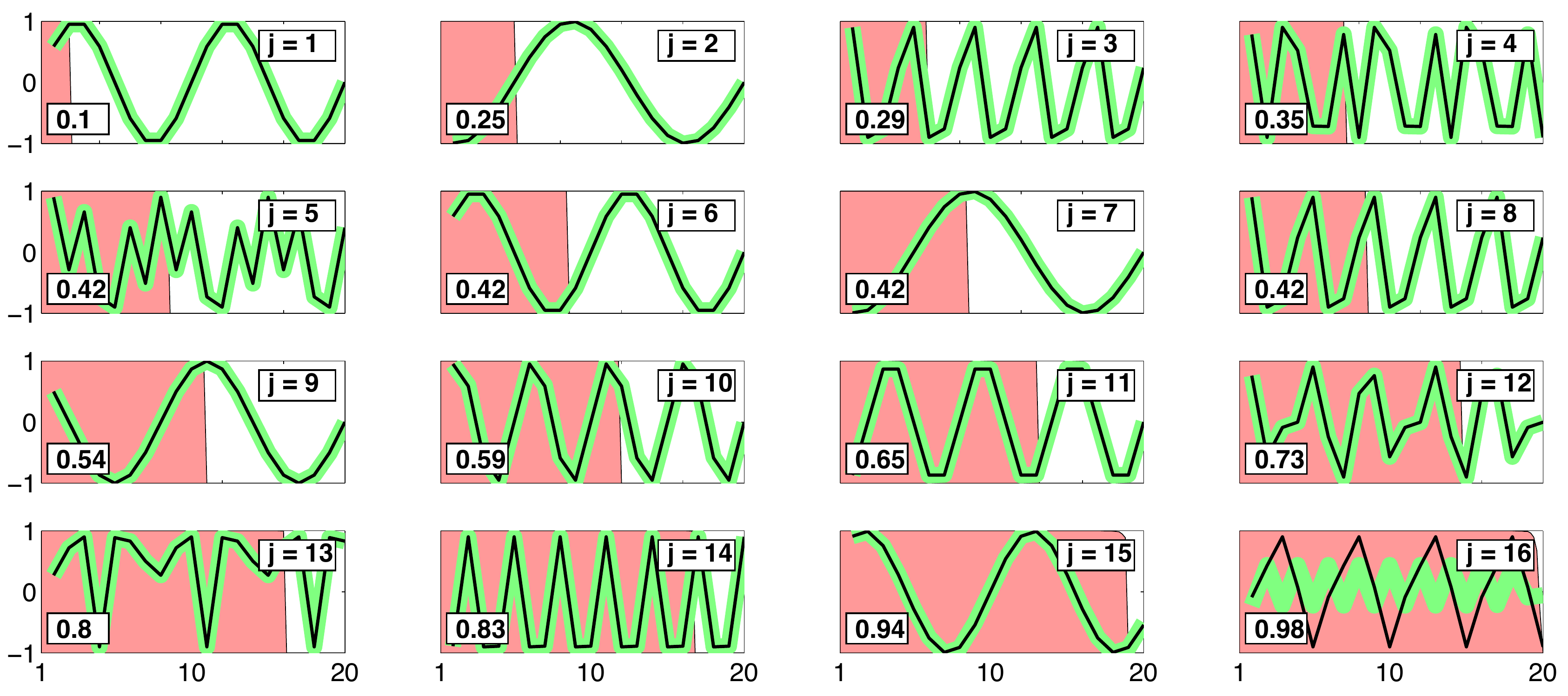}
\caption{Incremental pattern storing. Each panel shows a 20-timestep
  sample of the original training pattern $p^j$ (black line) overlaid
  on its conceptor-controlled reproduction (green line). The memory
  fraction used up until pattern $j$ is indicated by the panel fraction
  filled in red and its numerical value is printed in the left bottom corner
  of each panel.}
\label{fig8}
\end{figure}

 Demo: 16 patterns (some random
integer-periodic patterns, some sampled sines) were incrementally
stored in a 100-neuron RNN. After the last one was stored, the
re-generation quality of all of them was tested by
conceptor-controlled recall. Figure \ref{fig8} illustrates the outcome.
The memory space claimed at each stage is indicated by the red panel
filling; it is measured by a normalized size  of $A^n$ (the largest possible
such $A$ is the identity  conceptor with size 1). It can
be seen that the network successively ``fills up'', and is essentially
exhausted after pattern $p^{15}$: the sixteenth pattern cannot be
stored and its re-generation fails. Note that patterns 6--8 are
identical to patterns 1--3. The incremental storing procedure
automatically detects that nothing new has to be stored for $n$ = 6--8
and claims no
additional memory space.

Another practical use of Boolean operations is in dynamical pattern
classification. Again, with the aid of ``Boolean learning management''
a pattern classification system can be trained incrementally such that
after it has learnt to classify patterns $p^1,\ldots, p^n$, it can be
furthermore trained to recognize $p^{n+1}$ without re-visiting earlier
used training data. Furthermore the system can combine positive and
negative evidence, motto: ``this test pattern seems to be in class $j$
AND it seems NOT to be in any of the other classes $1$ OR $2$ OR
$\ldots$ OR $j-1$ OR $j+1$ OR $\ldots$. In the widely used Japanese
vowels benchmark (admittedly not super-difficult by today's
standards), a conceptor-enabled neural classifier based on an RNN with
only 10 neurons easily reached the performance level of involved
state-of-the-art classifiers at a very low computational cost
(learning time a fraction of a second on a standard notebook
computer). I note in passing that patterns need not be stored in this
application; the native network's response to test patterns yields
the basis for classification.

It is not always necessary to precompute conceptors and somehow store
or memorize them for later use. Instead, a network
$\mathcal{N}(p^1,\ldots,p^n)$ can re-generate the stored patterns
without precomputed conceptors by running a content-addressing
routine. To this end, at recall time the conceptor ultimately needed
for re-generating $p^j$ is initialized to the zero conceptor. The
network $\mathcal{N}(p^1,\ldots,p^n)$ is then driven by a short and
possibly corrupted \emph{cue} version of $p^j$. During this cueing
phase, the zero conceptor is quickly adapted to a preliminary version
$C^j_{\mbox{\scriptsize cue}}$ of $C^j$. When the cue signal expires, the
network run is continued in autonomous mode with the conceptor in the
loop. Its adaptation continues too. Since there is no external guide
it can adapt to, it adapts to ... itself! Using a human cognition
metaphor, this \emph{auto-adaptation} can be likened to the recognition
processing triggered by a brief stimulus, for instance when one gets a
passing glimpse of a face in a crowd and then in an ``recognition
afterglow'' consolidates this impression to the well-known face of a
friend. In terms of conceptor geometry, the ellipsoid shape of the
preliminary $C^j_{\mbox{\scriptsize cue}}$ is ``contrast-enhanced'' by
auto-adaptation: axes that are weak in $C^j_{\mbox{\scriptsize cue}}$
are further diminished and eventually are entirely suppressed, while
strong axes grow even stronger. Altogether in the auto-adaptation
phase $C^j_{\mbox{\scriptsize cue}}$ converges toward a
contrast-enhanced version $C^j_{\mbox{\scriptsize adapt}}$ of $C^j$.
This auto-adaptation dynamics has interesting and useful mathematical
properties.  In particular it is inherently robust against noise. In
the simulations reported in \cite{Jaeger14extended} it functions
reliably even in the presence of neural noise with signal-to-noise
ratios less than one. Furthermore, when the stored patterns
$p^1,\ldots,p^n$ are samples from a parametric family, the
content-addressed recall also functions when unstored members of this
family are used as cue (``class learning effect'').  For sufficiently
large $n$, the network $\mathcal{N}(p^1,\ldots,p^n)$ has implicitly
extracted the ``family law''.  A mathematical and numerical
investigation reveals that this class learning effect can be
interpreted as the creation of an approximate plane attractor under
the auto-adaptation dynamics in conceptor space. 

 \begin{figure}[htb]
\center
  \includegraphics[width=100mm]{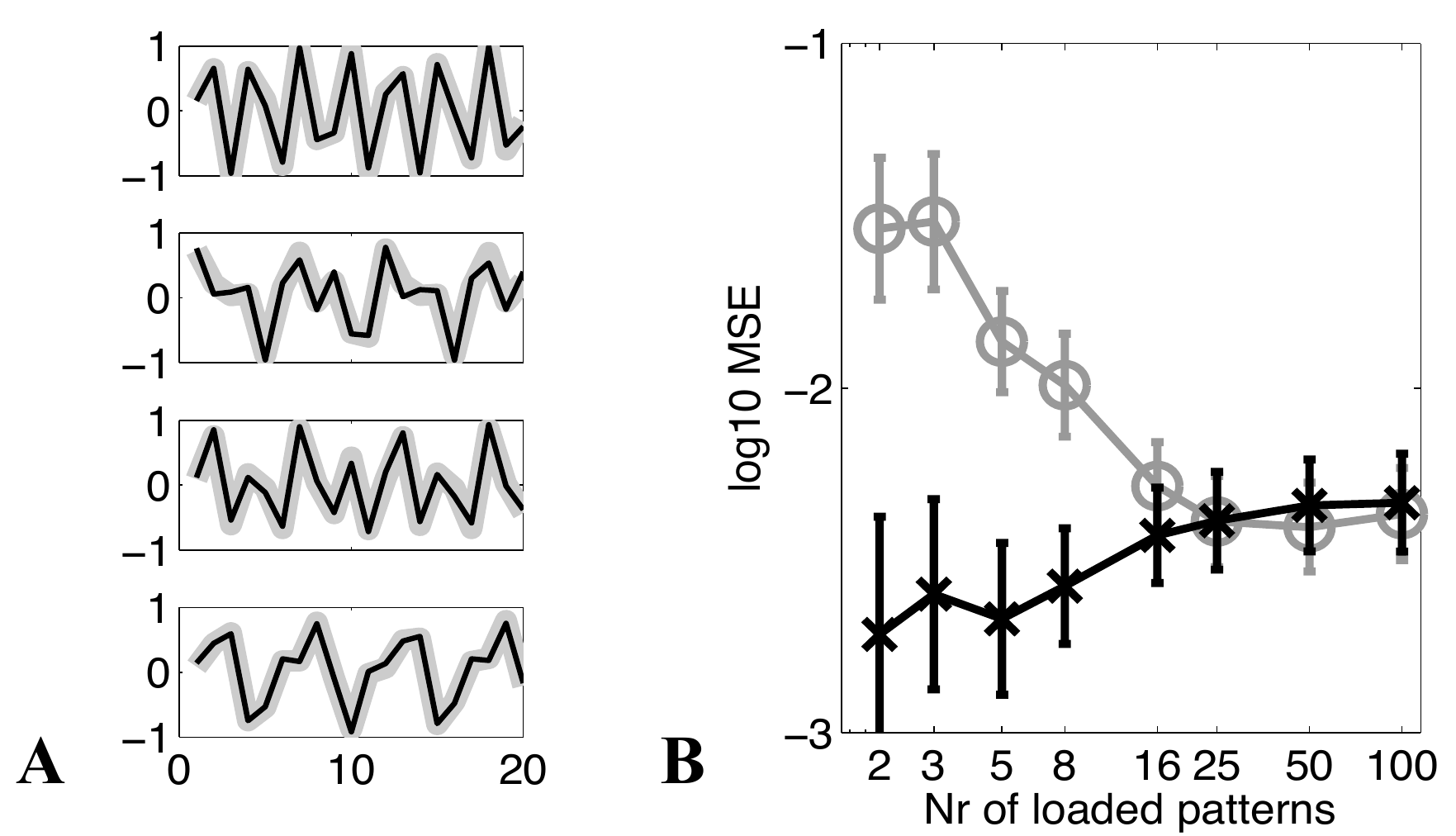}
\caption{Class-learning effect in content-addressable memory. {\bf A}
  Some instances from the 2-parametric pattern family (originals:
  black thin line, reconstructions: gray thick line). {\bf B}
  Reconstruction error (normalized root mean square error) at the end
  of auto-adaptation of stored patterns (black) and novel patterns
  (gray) versus the number of stored patterns. }
\label{fig9}
\end{figure}

Demo: Figure
\ref{fig9} shows the result of a simulation study where in separate
trials $n = 2, 3, 5, \ldots, 100$ patterns from a 2-parametric family
were stored. The networks $\mathcal{N}(p^1,\ldots,p^n)$ were tested
with cues that corresponded to stored patterns and with cues that came
from the pattern family but were not among the stored ones. For small
$n$, the stored patterns can be re-generated better than the novel
ones (a rote learning effect). When the number of stored
patterns exceeds a critical value, stored
and novel patterns are re-generated with equal accuracy and storing
even more patterns has no effect.

Altogether these  content-addressable neural memory systems can
be seen in many respects as a dynamical analog of  Hopfield networks,
the classical model of associative memories for static
patterns. 

\section{Conclusion}

Conceptors are a mathematical, computational, and  neural
realization of a simple pair of ideas:

\begin{itemize}
\item Processing modes of  an RNN can be 
characterized  by the geometries of the associated state
clouds.
  \item When the states of an RNN are filtered to remain within such a
specific state cloud, the associated processing mode is selected
and stabilized.
\end{itemize}

Implicit in this pair of ideas is the -- notrivial -- claim that a
single RNN may indeed host a diversity of processing modes. This is
the essence of conceptors:

\vspace{0.3cm}
\begin{center}
\fbox{Conceptors can control a multiplicity of processing modes of an RNN.}
\end{center}
\vspace{0.3cm}

Almost all examples in this article concerned a particular type of processing
mode, namely pattern generation. This bias is due to the circumstance
that conceptors were first conceived in the context of a research
project concerned with robot motor skills
(\href{http://http://www.amarsi-project.eu}{\tt
  www.amarsi-project.eu}). But a network governed by conceptors can be
employed in any of the sorts of tasks in which RNNs become engaged:
signal prediction, filtering, classification, control, etc. In
scenarios other than pattern generation it is often not necessary to
store patterns in the concerned RNN -- the storing procedure is not a
constitutive component of conceptor theory.

Conceptors offer rich possibilities to morph, combine, and adapt  an
RNN's processing modes through operations on conceptors: linear
mixing, logical operations and aperture adaptation. 

By virtue of logical combinations and conceptor abstraction, the
processing modes of an RNN can be seen as organized in a similar way
as concept hierarchies in AI formalisms. This has motivated to name
these operators ``conceptors''.

In my view these are the most noteworthy concrete innovations brought about by
conceptors so far:

\begin{itemize}
    \item they make it possible in the first place to characterize and
  govern a diversity of RNN processing modes,
\item they enable incremental pattern learning in RNNs with
an option to quantify and monitor claimed memory capacity,
\item they yield a model of an auto-associative memory for dynamical
patterns. 
\end{itemize}

A similarly noteworthy but more abstract and epistemological
innovation can be recognized in the firm link between nonlinear neural
dynamics and symbolic logic, established by the dual mathematical
nature of conceptors as neural state filters on the one hand and as
discrete objects of logical operations on the other. 

I am a machine learning researcher and this article undoubtedly
reflects limits of this perspective. In all examples that I presented
conceptors were derived from the simulated dynamics of simple
artificial RNNs.  But conceptors can be computed on the basis of any
sufficiently high-dimensional numerical timeseries. This indicates
usages of conceptors as a tool for data analysis and interpretation in
experimental disciplines. For instance, it sounds like an interesting
project for an empirical cognitive neuroscientist to (i) submit a
subject to a cognitive task which involves Boolean operations, (ii)
record high-dimensional brain activity of some sort, and (iii) check
to what extent conceptors derived from those recordings reflect the
Boolean relationships that are inherent in the task specification.  I
would actually not expect that this can be straightforwardly done with
any kind of raw signals. A more insightful question is to find out
\emph{which} brain data recorded from \emph{where} and transformed
\emph{how} do mirror logico-conceptual task characteristics.

I have carried out quite a number of diverse simulation experiments
with conceptors. Over and over again I was impressed by the
robustness of conceptor learning and operation against noise and
parameter variations. Furthermore, the basic algorithms are
computationally cheap. For a machine learning engineer like myself
they feel like really sturdy and practical enablers for building
versatile RNN-based information processing architectures. For
applications in  biological modeling (a field where I am no expert) I
would believe that robustness and cheapness are likewise relevant.

This appetizer article certainly does not qualify as a scientific
paper. A more serious account is provided by the technical report
\cite{Jaeger14extended} (about 200 pages).  Besides giving all the
formal definitions, algorithms, mathematical analyses, simulation
detail and references that are missing in the present article, it
expands on some further topics that I did not touch upon here.
Specifically, it explores the hairy issue of ``biological
plausibility'' and proposes (still rather abstract) neural circuits
which support conceptors and which only require local computations; it
analyzes conceptor auto-adaptation with tools from dynamical systems
theory; it specifies a formal logic which grounds symbolic conceptor
expressions in neural signal semantics; and it presents a
multi-functional hierarchical neural processing architecture wherein
higher processing levels inform and modulate lower processing levels
through conceptors.

My personal take on real brains and really good robots is that they
will never be fully explainable or designable on the basis of a single
unified theory.  I view conceptors as one further model mechanism 
which sheds  some more light on  some  aspects of system
integration in brains, animals, humans and robots.


\end{document}